\documentclass[9pt, conference]{IEEEtran}
\IEEEoverridecommandlockouts
\usepackage{cite}
\usepackage{amsmath,amssymb,amsfonts}
\usepackage{algorithmic}
\usepackage{graphicx}
\usepackage{textcomp}
\usepackage{multicol}
\usepackage{multirow}
\usepackage{xcolor}
\usepackage{pifont}
\usepackage{cite}
\usepackage{amsmath,amssymb,amsfonts}
\usepackage{algorithmic}
\usepackage{graphicx}
\usepackage{textcomp}
\usepackage{xcolor}
\usepackage[numbers]{natbib}
\usepackage{booktabs, multirow} % for borders and merged ranges
\usepackage{soul}% for underlines
\usepackage{changepage,threeparttable} % for wide tables

\usepackage{amsmath,tabularx}
\usepackage{makecell}
\usepackage{tablefootnote}
\usepackage{changepage,threeparttable}

\usepackage{hyperref}
\usepackage{times}
\usepackage{textcomp}
\usepackage{cite}

\def\BibTeX{{\rm B\kern-.05em{\sc i\kern-.025em b}\kern-.08em
    T\kern-.1667em\lower.7ex\hbox{E}\kern-.125emX}}
\begin{document}

\title{A Comparative Study of Discrete Speech Tokens for Semantic-Related Tasks with Large Language Models\\

}
% \author{
% \IEEEauthorblockN{
% Dingdong Wang\IEEEauthorrefmark{1},  Mingyu Cui\IEEEauthorrefmark{1}, Dongchao Yang\IEEEauthorrefmark{1}, Xueyuan Chen\IEEEauthorrefmark{1}, Helen Meng\IEEEauthorrefmark{1}}
% \IEEEauthorblockA{
% \IEEEauthorrefmark{1}The Chinese University of Hong Kong, Hong Kong SAR, China \\
%  \vspace{-0.5cm}
% }
% }

\author{
\IEEEauthorblockN{
Dingdong Wang\quad  Mingyu Cui\quad Dongchao Yang\quad Xueyuan Chen\quad Helen Meng}
\IEEEauthorblockA{
The Chinese University of Hong Kong, Hong Kong SAR, China \\
 \vspace{-0.5cm}
}
}
\maketitle

\label{0-abstract}

With the rise of Speech Large Language Models (Speech LLMs), there has been growing interest in discrete speech tokens for their ability to integrate with text-based tokens seamlessly. Compared to most studies that focus on continuous speech features, although discrete-token based LLMs have shown promising results on certain tasks, the performance gap between these two paradigms is rarely explored. In this paper, we present a fair and thorough comparison between discrete and continuous features across a variety of semantic-related tasks using a light-weight LLM (Qwen1.5-0.5B). Our findings reveal that continuous features generally outperform discrete tokens, particularly in tasks requiring fine-grained semantic understanding. Moreover, this study goes beyond surface-level comparison by identifying key factors behind the under-performance of discrete tokens, such as limited token granularity and inefficient information retention. To enhance the performance of discrete tokens, we explore potential aspects based on our analysis. We hope our results can offer new insights into the opportunities for advancing discrete speech tokens in Speech LLMs.

\section{Introduction}
Learning speech representations that are robust and effective is a key challenge in modern audio and speech processing systems. With the advancement of Speech Large Language Models (Speech LLMs), two main speech representations are employed as speech inputs: continuous features and discrete speech tokens.

Using speech continuous features is an intuitive approach~\cite{chu2023qwen,chen2023lauragpt,gong2023joint,wang2023blsp,tang2023salmonn,gong2023listen} for integrating speech signals with large language models (LLMs). In models like Qwen-Audio~\cite{chu2023qwen}, raw speech signals are transformed into high-dimensional embeddings using speech encoders such as Whisper Encoder~\cite{radford2023robust}, and are then adapted for LLM input through an adapter module. This method preserves the richness of the audio signal for downstream tasks and has demonstrated strong performance in speech understanding ability~\cite{chu2023qwen}. In this paradigm, speech is represented in continuous form while text is represented by discrete tokens. However, since autoregressive LLMs like GPT~\cite{liu2023gpt} and LLaMA~\cite{dubey2024llama} are naturally designed to work with discrete text tokens, this has inspired researchers to explore representing speech as sequences of discrete speech tokens.

Compared to continuous features, discrete tokens enable the seamless integration of speech signals with text-based tokens in LLMs. By compressing speech information, discrete tokens also enable more efficient transmission and storage. Owing to these advantages, several approaches have employed discrete tokens as input for Speech LLMs~\cite{zhang2023speechgpt,rubenstein2023audiopalm,wang2023viola,wang2024speechx}, such as Audiopalm~\cite{rubenstein2023audiopalm} and SpeechGPT~\cite{zhang2023speechgpt}. Both models utilize k-means clustering to discretize embeddings from speech representation models. Following the terminology from prior works~\cite{zhang2023speechtokenizer,borsos2023audiolm}, discrete tokens are categorized into two types: Acoustic tokens and Semantic tokens. Acoustic tokens (e.g., Soundstream~\cite{zeghidour2021soundstream}, Encodec~\cite{defossez2022high}) are obtained using compression-based methods, which rely on encoder-decoder architectures with residual vector quantization (RVQ)~\cite{zeghidour2021soundstream}. Semantic tokens involve applying clustering algorithms such as K-means to extract features from SSL models, using the cluster indices as discrete representations.

Previous studies show that Speech LLMs employing continuous features excel in semantic-related tasks, demonstrating strong speech understanding abilities. However, it is still unclear whether discrete tokens can achieve similar levels of speech comprehension. While the prior study~\cite{chang2024exploring} found discrete tokens outperform continuous features on many datasets, these conclusions may not consistently apply to Speech LLMs. Additionally, existing research on discrete tokens primarily focuses only on certain specific tasks, such as TTS ~\cite{wang2023neural,yang2024towards}, ASR~\cite{chang2023exploration,yang2024towards}, and speech translation~\cite{zhang2023dub,zhang2023dub,kim2023many}, these works have not been thoroughly benchmarked. Without comprehensive evaluation, it remains uncertain whether the conclusions from these focused studies can be generalized.

To address the gap, we aim to conduct a comprehensive comparative study of discrete speech tokens versus continuous features in LLM-based semantic tasks. Specifically, we employ the K-means clustering technique to generate semantic tokens from self-supervised learning (SSL) models~\cite{hsu2021hubert,chen2022wavlm}. This method is chosen for its simplicity, efficiency, and ability to capture semantic information without extra training. On a wide variety of tasks and benchmarks, our findings suggest that continuous features generally perform better in most tasks. Then, we conduct further exploration to investigate the potential factors behind the gap, uncovering the discrete tokens' potential for improvement and scalability. Our key contributions are as follows:
\begin{itemize}
    \item 
    % A comparative analysis of continuous features and discrete tokens is conducted across five semantic tasks (speech recoginition, phoneme recognition, speech translation, intent classification, and keyword spotting). 
    To the best of our knowledge, this is the first comprehensive comparison of LLM-based continuous features and discrete tokens on multiple semantic-related tasks (automatic speech recognition, phoneme recognition, speech translation, intent classification, and keyword spotting).
    % \item We assess other attributes (i.e. the training efficiency and data size scalability) of both continuous features and discrete speech tokens.
    \item We comprehensively evaluate the performance of both continuous features and discrete speech tokens in terms of training efficiency and data size scalability, providing crucial insights for speech processing model selection.
    % \item We investigate the potential factors behind the performance gap between continuous features and discrete tokens. 
    \item We conduct an in-depth analysis to uncover the key factors contributing to the performance gap between continuous features and discrete tokens.
\end{itemize}
\section{Pipeline Design}

\begin{figure}[!t]
    \centering
    \includegraphics[width=0.45\textwidth]{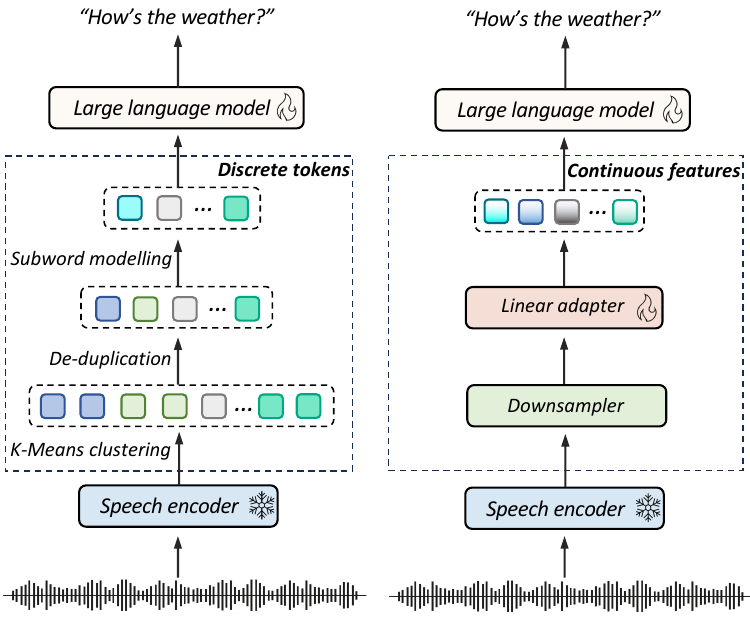}
    \caption{Architectures of two approaches for integrating speech into Large Language Models (LLMs): 
 \textbf{discrete token-based encoding} versus \textbf{continuous feature processing}.}
    \label{fig:arch}
\end{figure}

To systematically compare the performance of continuous features and discrete tokens in speech tasks, we adopted two widely used speech input processing pipelines, as illustrated in Figure~\ref{fig:arch}: (a) a discrete token pipeline consisting of K-means clustering, de-duplication, and subword modelling, and (b) a continuous features pipeline incorporating a downsampler and a linear adapter module. These approaches align with recent research \cite{chu2023qwen,gong2023joint,wang2023blsp,zhang2023speechgpt,shon2024discreteslu,dekel2024exploring}.

\subsection{Processing Continuous Features}

In our approach, we utilize self-supervised learning (SSL) models (HuBERT-Large~\cite{hsu2021hubert} and WavLM-Large~\cite{chen2022wavlm}) to generate continuous features. The processing pipeline involves two steps: (1) \textbf{Downsampling}, where the feature sequence \( H = \{h_1, h_2, \dots, h_T\} \) is reduced by concatenating every \( k \) consecutive frame, producing \( Z^S = \{z_1^S, \dots, z_N^S\} \) with \( N = T/k \), lowering computational complexity while retaining key information. (2) \textbf{Linear Adapter}: After downsampling, we use a single linear adapter
layer to map the embeddings to the LLM’s hidden size.

\subsection{Speech Discretization}
For discrete tokens, we utilize the same speech encoders. We first generate a sequence of high-dimensional feature vectors \( H = \{h_1, h_2, \dots, h_T\} \) from the audio waveform using an SSL model. Second, we apply \textbf{K-Means Clustering} to obtain discrete tokens \( Z = \{z_1, z_2, \dots, z_T\} \). These tokens now exist as discrete symbols that can be processed like text symbols in NLP, making it possible to use traditional NLP techniques.

After clustering, the discrete token sequence may contain redundant consecutive information. Thus we apply \textbf{De-duplication} to merge consecutive identical tokens to reduce redundancy, and finally use \textbf{Byte-Pair Encoding (BPE)} to enhance input tokens by combining frequent subsequences into shorter meta tokens \( Z' = \{z'_1, z'_2, \dots, z'_{T'}\} \), where \( T' < T \). Mathematically, the length reduction ratio \( \frac{T'}{T} \) typically ranges from 30\% to 60\%, depending on the k-means clustering granularity and the BPE vocabulary size.

\begin{table}[!t]\centering
\footnotesize
\caption{\parbox{0.95\linewidth}{\centering Effect of different settings on ASR with LibriSpeech 960h.}}
\label{tab:km-settings}
\vspace{-0.2cm}
\renewcommand{\arraystretch}{0.8}
\begin{tabular}{l|cc|cc}
\toprule
\multirow{2}{*}{\textit{SSL model}} & \multicolumn{2}{c}{\textit{Manipulation}} & \multicolumn{2}{|c}{WER $\downarrow$} \\ 
\cmidrule{2-5}
& K-Means & BPE size &test-clean & test-other \\
\midrule
\multirow{6}{*}{Hubert-Large} & k=1000 & -& 7.48 & 12.82\\
& k=2000 & -& 5.02 & 10.55\\
& k=3000 & -& 5.01 & 10.51\\
& k=2000 & 4000& 4.99 & 10.78\\
& k=2000 & 6000& \textbf{4.56} & \textbf{9.79}\\
& k=2000 & 8000& 5.04 & 11.20\\
\midrule
\multirow{6}{*}{WavLM-large} & k=1000 & -& 5.33 & 10.54\\
& k=2000 & -& 5.04 & 10.11\\
& k=3000 & -& 5.03 & 10.34\\
& k=2000 & 4000& 4.88 & 10.62\\
& k=2000 & 6000& 4.72 & \textbf{10.45}\\
& k=2000 & 8000& \textbf{4.62} & 10.82\\
\bottomrule
\end{tabular}
\end{table}

\subsection{Instruction-Tuning with LLM}

We perform \textbf{instruction-tuning} to adapt the discrete and continuous speech inputs for various tasks, enabling the LLM to follow specific instructions.

\begin{enumerate}
    \item \textbf{Tasks and Instructions Prompts}:
    \begin{itemize}
        \item \textbf{Automatic Speech Recognition (ASR)}: \textit{"Identify the text corresponding to the speech."}
        \item \textbf{Phoneme Recognition (PR)}: \textit{"Identify the phonemes corresponding to the speech."}
        \item \textbf{Keyword Spotting (KS)}: \textit{"Identify the keyword corresponding to the speech."}
        \item \textbf{Emotion Recognition (ER)}: \textit{"Identify the emotion corresponding to the speech."}
        \item \textbf{Spoken Intent Classification (IC)}: \textit{"Identify the intention corresponding to the speech."}
        \item \textbf{Speech Translation (ST)}: \textit{"Translate the English speech into Chinese (or German) text."}
    \end{itemize}
    
    \item \textbf{LLM Decoder}: We mainly use the light-weight \textit{Qwen1.5-0.5B}~\cite{bai2023qwen} model as the decoder to explore the performance of discrete tokens and continuous features across various tasks. Additionally, to investigate how discrete tokens perform with different scales of LLM decoders, we also utilize the \textit{LLaMA3.1-8B} model~\cite{dubey2024llama}. This allows us to analyze the impact of LLM decoder size on the effectiveness of discrete speech tokens. 
    
\end{enumerate}

\begin{table*}[t]
\centering
\caption{\parbox{0.95\linewidth}{\centering Comparison of discrete and continuous speech tokens on various tasks based on Qwen-1.5-0.5B model. Discrete tokens use K-means (2000 clusters) with BPE size 6000 for all tasks.}}
\label{tab:main-qwen}
\footnotesize
\vspace{-0.2cm}
\renewcommand{\arraystretch}{1.3}  % 使表格行高变大
\setlength{\tabcolsep}{4pt}
\begin{tabular}{ll|c|c|c|c|c|c|c}
\toprule

\multirow{2}{*}{\textit{SSL model}} & \multirow{2}{*}{\textit{Token type}} & \multicolumn{2}{c}{ASR (WER$\downarrow$)} & \multicolumn{1}{|c|}{PR (PER$\downarrow$)} & \multicolumn{1}{c|}{ST (BLEU$\uparrow$)} & \multicolumn{1}{c|}{KS (ACC$\uparrow$)} & \multicolumn{1}{c|}{IC (ACC$\uparrow$)} & \multicolumn{1}{c}{ER (ACC$\uparrow$)} \\ \cmidrule{3-4} \cmidrule{5-9}

& & \makecell{LibriSpeech \\ (test-clean$\vert$other)} & \makecell{Gigaspeech-M\\(test)} & \makecell{Libri-100\\(test-clean)} & \makecell{GigaST\\(En-Zh$\vert$En-De)} & \makecell{Speech Commands\\(test$\vert$val)} & \makecell{SLURP\\(test)} & \makecell{IEMOCAP\\(test)} \\ \midrule

HuBERT-Large & \multirow{2}{*}{Discrete} & 4.56 / 9.79 & 19.40 & 9.69 & 22.75 / 20.14 & 93.70 / 93.85 & 57.04 & 38.65  \\
WavLM-Large &  & 4.72 / 10.45 & 16.34 & \textbf{9.64} & 24.62 / 21.22 & 92.87 / 92.45 & 59.96 & 37.98 \\ \hline

HuBERT-Large & \multirow{2}{*}{Continuous} & 4.91 / 6.43 & 17.45 & 12.84 & 26.63 / 25.42 & 95.38 / 95.70 & 76.84 & 56.72 \\
WavLM-Large &  & \textbf{2.92} / \textbf{4.61} & \textbf{13.96} & 12.62 & \textbf{29.44} / \textbf{28.12} & \textbf{97.76} / \textbf{97.36} & \textbf{81.35} & \textbf{59.45} \\

\bottomrule
\end{tabular}
% \\ \scriptsize{feel free to edit here}
\end{table*}
\begin{table*}[t]
\centering
\caption{\parbox{0.95\linewidth}{\centering Performance comparison of discrete speech tokens using Llama 3.1-8B. Same settings as in Qwen 1.5-0.5B.}}
\label{tab:main-llama}
\footnotesize
\vspace{-0.2cm}
\renewcommand{\arraystretch}{1.3}  % Adjust row height
\setlength{\tabcolsep}{4pt}  % Adjust column spacing
\begin{tabular}{ll|c|c|c|c|c|c|c}
\toprule

\multirow{2}{*}{\textit{SSL model}} & \multirow{2}{*}{\textit{Token type}} & \multicolumn{2}{c}{ASR (WER$\downarrow$)} & \multicolumn{1}{|c|}{PR (PER$\downarrow$)} & \multicolumn{1}{c|}{ST (BLEU$\uparrow$)} & \multicolumn{1}{c|}{KS (ACC$\uparrow$)} & \multicolumn{1}{c|}{IC (ACC$\uparrow$)} & \multicolumn{1}{c}{ER (ACC$\uparrow$)} \\ \cmidrule{3-4} \cmidrule{5-9}

& & \makecell{LibriSpeech \\ (test-clean$\vert$other)} & \makecell{Gigaspeech-M\\(test)} & \makecell{Libri-100\\(test-clean)} & \makecell{GigaST\\(En-Zh$\vert$En-De)} & \makecell{Speech Commands\\(test$\vert$ val)} & \makecell{SLURP\\(test)} & \makecell{IEMOCAP\\(test)} \\ \midrule

HuBERT-Large & \multirow{2}{*}{Discrete} & \textbf{2.56} / \textbf{6.49} & \textbf{12.859} & 7.85 & 26.32 / 25.24 & 96.75 / 96.69 & 63.44 & \textbf{39.84}  \\
WavLM-Large &  & 2.96 / 7.48 & 13.35 & \textbf{7.02} & \textbf{28.62} / \textbf{26.87} & \textbf{97.92} / \textbf{98.17} & \textbf{66.96} & 36.12 \\

\bottomrule
\end{tabular}
\end{table*}

\label{3-experiment}
\section{Experiments}
\subsection{Experimental Setup}

To investigate the performance of discrete tokens compared to continuous features across various semantic tasks, we conduct experiments using a range of datasets: LibriSpeech~\cite{panayotov2015librispeech} and GigaSpeech M-size~\cite{chen2021gigaspeech} for \textbf{Automatic Speech Recognition (ASR)}, LibriSpeech 100-hour~\cite{panayotov2015librispeech} for Phoneme Recognition (PR), Speech Commands-v2~\cite{warden2018speech} for \textbf{Keyword Spotting (KS)}, SLURP~\cite{bastianelli2020slurp} for \textbf{Spoken Intent Classification (IC)}, and GigaST~\cite{ye2022gigast} for \textbf{Speech Translation (ST)}. To evaluate whether the semantic discrete tokens capture paralinguistic emotion information, we also included an \textbf{Emotion Recognition (ER)} task using the IEMOCAP dataset~\cite{busso2008iemocap}. Due to resource constraints, we mainly focused on the Qwen1.5-0.5B model, with further discrete token experiments conducted on the larger LLaMA3.1-8B model. For Qwen1.5-0.5B, full-parameter fine-tuning was used. For LLaMA3.1-8B, we applied LoRA~\cite{hu2021lora} for efficiency by injecting adapters (rank=8 and \( \alpha = 16 \)) into the projection layers for keys and queries in all self-attention layers. 

For a fair comparison, we consistently extracted features from the final layer of both HuBERT-Large~\cite{hsu2021hubert} and WavLM-Large~\cite{chen2022wavlm} at a 16,000 Hz sampling rate for both discrete and continuous tokens. We chose K-means with 2000 centroids and 6000 BPE vocabulary size for generating discrete tokens across all tasks, based on preliminary experiments. For continuous tokens, we applied a downsampling rate of \( \alpha = 2 \) and a single-layer linear adapter to project the embeddings to the LLM input space. All experiments were conducted under identical settings, with a learning rate of \( 1 \times 10^{-5} \), batch size of 32, using the AdamW optimizer with a weight decay of \( 10^{-2} \). 

\subsection{Discrete Tokens ASR Results}

We first evaluated the effect of varying k-means centroids and BPE vocabulary size on discrete tokens ASR performance on the LibriSpeech 960-hour dataset. As shown in Table~\ref{tab:km-settings}, increasing the number of centroids generally improved performance. BPE-based subword modelling can further improve WER results by reducing token sequence length while preserving key semantic information. Notably, the combination of \( k = 2000 \) centroids and 6000 BPE vocabulary size achieved a balanced trade-off between performance gains and computational efficiency. We fix this setting for all datasets to maintain discrete tokens consistency across tasks.

\subsection{Comparison Results}

Our results using the Qwen1.5-0.5B model~\cite{bai2023qwen} (as shown in Table~\ref{tab:main-qwen}) indicate that continuous features generally outperform discrete tokens in most downstream tasks, except phoneme recognition. WavLM-Large outperforms HuBERT-Large concerning to the continuous features. While K-means-based discrete tokens capture semantics, they struggle with emotion information, indicating continuous tokens better preserve fine-grained details crucial for sentiment tasks. Moreover, while HuBERT-Large performs similarly with discrete and continuous features on the LibriSpeech test-clean dataset, continuous features outperform on the noisier test-other dataset. For complex tasks like speech translation and intent classification, the performance gap increases, showing the limitations of discrete tokens in capturing deeper context and meaning.

\subsection{Training Efficiency of Different Features}

\begin{figure}[!t]
    \centering
    \includegraphics[width=0.45\textwidth]{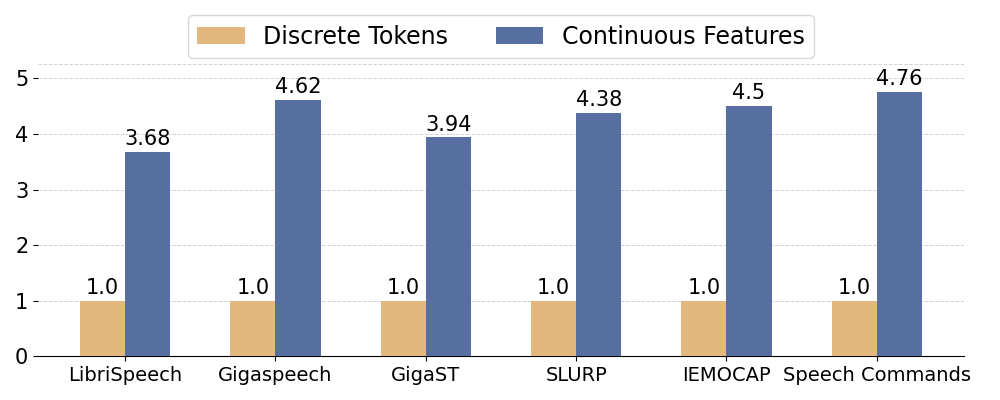}
    \caption{Total training time until convergence for discrete tokens and continuous features, with discrete token training time normalized to 1 for all datasets.}
    \label{fig:training_time}
\end{figure}

Figure~\ref{fig:training_time} illustrates the total training time required until convergence for both discrete tokens and continuous features across several datasets. We normalize the training time of discrete tokens to 1 for comparison. Based on our experiments across all datasets, discrete tokens converge within 4$\sim$5 epochs, while continuous features require 14$\sim$15 epochs. This leads to a significant reduction in total training time for discrete tokens, ranging from approximately 21\% to 27\% of the time required for continuous features across the datasets. The efficiency gain is due to the compact representation of discrete tokens. This compression not only accelerates the training process but also lowers the demand for computational resources.

\label{4-discussion}
\section{Discussion}

We observed that continuous features generally outperform discrete ones in most tasks, raising questions about the factors behind this performance gap. Thus, we aim to explore these differences and investigate potential improvements for discrete tokens in future applications.

\begin{table}[ht]
\centering
\caption{Data size (bit) comparison of a T-second utterance among: 1) SSL-based features using HuBERT-Large; 2) discrete tokens in 13-bit with 25 frames/sec. Reduction Ratio is calculated compared to the previous row based on LibriSpeech-100h.}
\renewcommand{\arraystretch}{0.8}
\label{tab:data-size}
\vspace{-0.2cm}
\setlength{\tabcolsep}{4pt}  % Adjust column separation
\begin{tabular}{l|l|l}
\toprule
Method       & Data Size (bits)     & Reduction Ratio \\ \midrule
SSL features  & $32 \times 1024 \times 25 \times T$   & \quad --\\
K-Means          & $13 \times 50 \times T$   & 99.90\% \\
+ De-duplication      & $13 \times 36.47 \times T$ & 27.06\%   \\
+ BPE            & $13 \times 20.24 \times T$   & 44.56\% \\ \bottomrule
\end{tabular}
\end{table}

\subsection{Potential Factors Affecting Discrete Speech Tokens Performance}

\textbf{1) Impact of Bitrate:}

A key factor contributing to the performance gap between discrete tokens and continuous features in LLMs is the associated bitrate. The bitrate \(R\), defined as \( R = \log_2 V \cdot C \cdot R_s \), where \( V \) is the vocabulary size, \( C \) is the number of codebooks, and \( R_s \) is the rate of codes per second, determines how well the codebook size retains information. Smaller codebooks result in lower bitrates, which may lose subtle acoustic nuances, while higher bitrates in continuous features preserve richer speech details.

Specifically, table~\ref{tab:data-size} compares the data size needed to represent a \( T \)-second speech utterance using continuous SSL features and compressed discrete tokens. Continuous features from HuBERT-Large require \( 32 \times 1024 \times 25 \times T \) bits (same as WavLM-Large), where 32 is the bit depth, 1024 is the feature dimensionality, and 25 is the frame rate. In contrast, K-means compresses the data to only \( 13 \times 50 \times T \) bits, with 13-bit (maximum 6000 clusters $\approx$ 13-bit) and 50 frames per second. It reduces the data size to less than 1/1000 of the continuous features. Further compression can be achieved through de-duplication and BPE subword modelling. The reduction ratio calculation is based on the LibriSpeech 100-hour dataset.

Table~\ref{tab:km-settings} highlights the effect of different bitrates levels on the performance of discrete tokens. Increasing the K-means codebook size from 1000 to 2000 reduces ASR word error rates (WER), but further increases yield diminishing returns. We also find that the introduction of Byte-Pair Encoding (BPE) improves performance by reducing token sequence length, demonstrating that achieving a balance between token size and compression efficiency can mitigate performance loss.

\textbf{2) Impact of Discrete Token Distribution Imbalance:}

\begin{figure}[!t]
    \centering
    \includegraphics[width=0.45\textwidth]{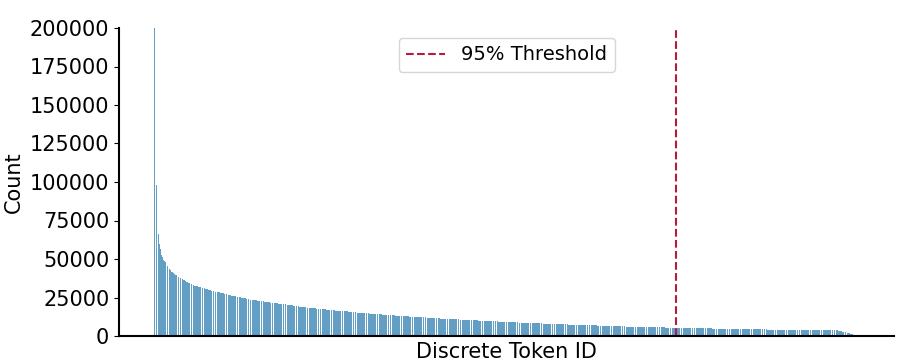}
    \caption{Frequency distribution of discrete tokens with a codebook size of 6000, based on the Gigaspeech M-size corpus. The red line indicates the 95\% cumulative frequency threshold.}
    \label{fig:imbalance}
\end{figure}

The graph~\ref{fig:imbalance} shows the frequency distribution of discrete tokens with a codebook size of 6000 (Optimal settings based on ASR experiments), based on the Gigaspeech M-size dataset. Around 20\% of distinct tokens account for only 5\% of total occurrences (similar pattern consistent across other datasets). This imbalance in token usage highlights a key issue with discrete tokenization: much of the codebook is underutilized, leading to inefficient representation. This uneven distribution introduces noise during LLM training, as the model may pay more attention to more frequent tokens while failing to adequately learn less common ones, resulting in poor generalization for speech segments represented by underutilized tokens. In contrast, continuous embeddings provide a more even and detailed representation, avoiding sparse token utilization and capturing subtle linguistic and acoustic features more effectively.

\textbf{3) Robustness of Speech Tokenizer:}

The results in Table~\ref{tab:strategy} highlight the robustness limitations of semantic discrete tokens, particularly in handling the challenges of the SLURP dataset, which includes noisy data, varied pronunciations (e.g., Indian English), and a mix of close and distant microphone recordings. These complexities make it difficult for discrete tokens to capture semantic content effectively. The table shows that first pretraining with SLURP transcriptions for ASR ("+SLURP ASR") significantly enhances SLURP intent classification (IC) performance for both HuBERT-Large and WavLM-Large, increasing accuracy by nearly 10\%. Adding more ASR data from LibriSpeech can further improve the IC accuracy. This highlights opportunities for further refinement in speech tokenizer design and the potential to enhance their robustness in handling complex tasks.

\begin{table}[!t]
\centering
\footnotesize
\caption{\parbox{1\linewidth}{\centering Intent classification (IC) performance with different training strategies. Baseline uses SLURP data for IC. "+SLURP ASR" adds ASR pretraining on SLURP transcripts, "+Libri-100" includes additional LibriSpeech 100-hour data.}}
\label{tab:strategy}
\vspace{-0.2cm}
\renewcommand{\arraystretch}{0.8}
\setlength{\tabcolsep}{5pt}  % Adjust column spacing
\begin{tabular}{lc|>{\centering\arraybackslash}p{2cm}}  % Use p{} to adjust only the ACC column width and center contents
\toprule
\multirow{2}{*}{\textit{SSL model}} & \multirow{2}{*}{\textit{Training strategy}} & \multicolumn{1}{c}{ACC $\uparrow$} \\ 
\cmidrule{3-3}
& & test \\  % Ensure "test" is centered
\midrule
HuBERT-Large & Baseline & 57.04 \\
& \hspace{0.5cm}+SLURP ASR & 66.36 \\
& \hspace{1cm}+Libri-100 & 67.18 \\
\midrule
WavLM-Large & Baseline & 59.96 \\
& \hspace{0.5cm}+SLURP ASR & 71.24 \\
& \hspace{1cm}+Libri-100 & 71.96 \\
\bottomrule
\end{tabular}
\end{table}

\subsection{Is There Room for Discrete Tokens to Improve?}

Another crucial question arises: is there room for improvement in discrete tokens performance? Thus we turn our focus to exploring potential aspects that could enhance their effectiveness.

\textbf{a) Explore Discrete Tokens on larger-scale LLM:}

Experiments with larger LLM (LLaMA3.1-8B model) shown in Table~\ref{tab:main-llama}, reveal that discrete tokens show notable improvement when integrated into larger decoder, particularly in ASR tasks. These experiments demonstrate that the performance of discrete tokens on the Qwen1.5-0.5B model is not the upper limit, as their capabilities expand significantly when used with larger LLM.

\textbf{b) Layer Analysis:}

\begin{figure}[!t]
    \centering
    \includegraphics[width=0.45\textwidth]{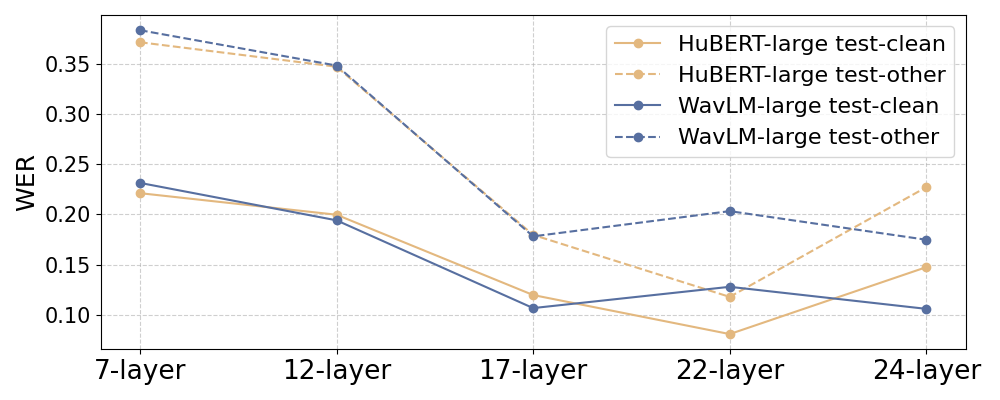}
    \caption{WER results of different layers based on Qwen1.5-0.5B with LibriSpeech 100-hour dataset.}
    \label{fig:wer_per_layer}
\end{figure}

SSL models capture varying levels of information across layers, with some layers better suited for specific tasks. Here, we analyze the impact of SSL models different layers on ASR performance. As shown in Figure~\ref{fig:wer_per_layer}, we evaluate using the LibriSpeech 100-hour dataset. Layer selection for ASR was guided by findings from prior work~\cite{chen2022wavlm}, revealing significant performance variation across layers. For WavLM-Large, the 17th layer achieves the lowest WER, while the 22nd layer performs best for HuBERT-Large.

\label{4-conclusion}
\section{Conclusion}

This work provides a comprehensive comparison of discrete and continuous speech features, finding that continuous features generally perform better, especially in tasks requiring detailed semantic understanding. By exploring the underlying factors, such as limited granularity and inefficient information retention, we propose several potential ways to enhance the performance of discrete tokens. In future work, we aim to develop a more robust and unified speech tokenizer for discretization.

\section{Acknowledgements}

Sincere thanks to Dr. Chen Xiao, Dr. Tan Daxin and Dr. Chu Ruihang for their invaluable guidance and support throughout this research. Their expertise and insightful feedback have greatly helped this work.

% \newpage
\bibliographystyle{IEEEtran}
\bibliography{IEEEfull,mybib}

\end{document}